# AN APPLICATION OF NON-MONOTONIC PROBABILISTIC REASONING TO AIR FORCE THREAT CORRELATION[1]

Kathryn B. Laskey and Marvin S. Cohen
Decision Science Consortium, Inc.
7700 Leesburg Pike, Suite 421
Falls Church, Virginia 22043

1. **Overview of the System.** Current approaches to expert systems' reasoning under uncertainty fail to capture the iterative revision process characteristic of intelligent human reasoning. This paper reports on a system, called the Non-monotonic Probabilist, or NMP (Cohen, et al., 1985). When its inferences result in substantial conflict, NMP examines and revises the assumptions underlying the inferences until conflict is reduced to acceptable levels. NMP has been implemented in a demonstration computer-based system, described below, which supports threat correlation and in-flight route replanning by Air Force pilots.

The NMP system uses a belief function representation of uncertainty (Shafer, 1976), embedded within a process of non-monotonic reasoning. Belief functions were chosen because, unlike Bayesian probabilities, they take explicit account of the *completeness* of an evidentiary argument, and they provide a natural measure of the conflict between two or more arguments. The non-monotonic reasoning process is based on an internal representation (Figure 1) for the structure of an evidential argument. This structure is originally due to Toulmin et al. (1984). The conclusion of an argument, or claim, is supported by grounds, or evidence. In NMP, the conclusion is not a definite hypothesis, but rather a belief function over the set of possible hypotheses (corresponding to Toulmin's conception of probability as a modal qualifier of a claim). The link between evidence and conclusion is provided by a rule for deriving a belief function from the evidence; this rule is backed by a deeper theoretical or causal model justifying the construction of the belief function. Toulmin allows for *possible rebuttals* which weaken the link between the evidence and the conclusion by asserting conditions under which the rule is invalidated. Such invalidation is represented naturally within the theory of belief functions by *discounting* of the output belief function.

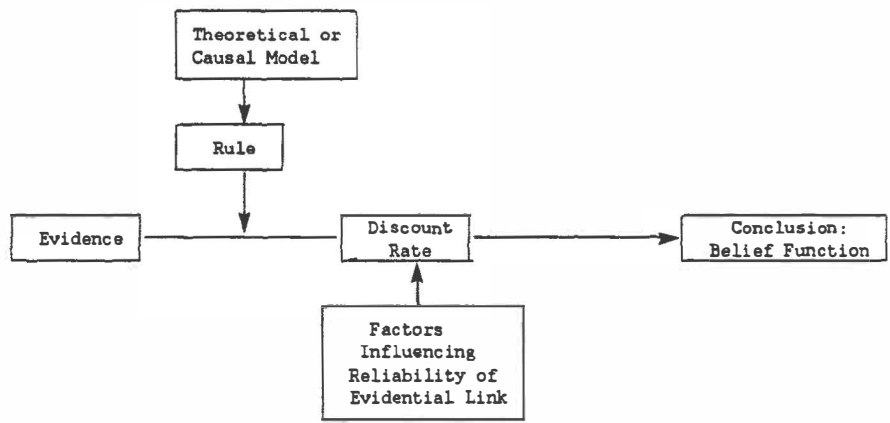

Figure 1: Structure of an Argument in NMP

----

This framework resembles non-monotonic logic (e.g., Doyle, 1979) in that conclusions are accepted *unless* other propositions (members of the outlist; rebuttals) turn out to be true. Two important differences are: (1) NMP is based on a highly differentiated knowledge structure, unlike the essentially homogeneous knowledge structure of non-monotonic logic; and (2) NMP provides an explicit measure of the uncertainty in an argument. The knowledge structure of NMP gives the system access both to the theoretical basis for deriving a belief function and to the factors which could discredit the link between evidence and conclusion. As a consequence, the system has the potential for "reaching inside" an argument in the process of conflict resolution, and identifying or adjusting the assessments that led to the conflict.

2. Combining Evidence and Conflict Resolution. The inference mechanism of NMP is based on Dempster's Rule for combining belief functions. The validity of Dempster's Rule depends on independence of the evidence on which the belief functions are based. Every attempt should be made to construct the system in such a way that independence holds. If this is not feasible, a more complex analysis is possible (Shafer, in press), but if the departure from independence is not too serious, one may judge that the improvement in results is not worth the substantial increase in complexity. In any case, the current version of NMP is based on Dempster's Rule.

When combined belief functions are in basic agreement, the inference procedure ends with the application of Dempster's Rule. In contrast, the presence of conflict (i.e. significant belief assigned to mutually exclusive conclusions), indicates a flaw in one of the component arguments, and routine statistical aggregation would be inappropriate. In such situations, NMP, like a human expert, examines its assumptions to determine the cause of the conflict. This process of conflict resolution involves the following steps. (1) The first step is to seek information that may discredit one of the component arguments. This may result in increasing belief in the presence of a *discrediting factor* which would discredit one of the arguments. If so, the discount rate of the "culprit" argument is increased, thereby reducing the conflict. (2) If significant conflict remains, the system may seek additional independent information relating to the conclusion. This typically will *increase* conflict, but may result in insight into the cause of the conflict. Going back to step 1, the system uses this new information to reprioritize its investigation of discrediting factors. (3) If conflict is still unresolved, the system may explore modifications in the theoretical base for one of the conflicting arguments. Unlike the investigation of discrediting factors, this step may involve a change in the *direction* in which the argument points. (4) Finally, if significant conflict remains, the system discounts all component arguments, using a formula that takes into account each argument's contribution to the conflict (one might also include a measure of the firmness, or resistance to discounting, of each argument).

The discrediting factors are represented by an internal knowledge structure. Each discrediting factor has associated with it (1) an *initial* belief function representing initial belief in factor presence (no information is represented as a vacuous belief function); (2) one or more *tests* for factor presence (representing information search options); (3) a set of *possible outcomes* for each test, each represented by a belief function for factor presence; and (4) a *cost* associated with performing each test.

The discount rate for each belief function is equal to the amount of belief directly committed to the *absence* of one or more discrediting factors. This amounts to a default assumption that all uncommitted belief is assigned to factor absence (Cohen, Laskey, and Ulvila, 1986). The belief function is dis-

160

counted by multiplying belief in all focal elements by 1 minus the discount rate, and increasing belief in the universal set so that beliefs sum to one. If the initial belief function is non-vacuous, discounting occurs prior to application of Dempster's Rule. When conflict occurs above a threshold level, our prototype system implements steps 1 and 4 of the conflict resolution procedure as follows. First, the system decides on a "culprit" discrediting factor and a test to perform by weighing its cost against its potential for conflict reduction. If no test is worth performing, step 4, or across-the-board discounting, is invoked. Otherwise, the system performs the chosen test, updates the belief function over factor presence, computes a new discount rate for the conclusion belief function, and recombines the new discounted belief function with the others. This process is repeated until conflict reaches an acceptable level.

This conflict resolution procedure is non-monotonic in character, and posesses important parallels to Doyle's (1979) non-monotonic logic. In a strict Shaferian analysis, the *input* belief functions remain fixed, and adding more evidence always *increases* the reliability of the conclusion (in the sense that less belief is committed to the universal set). Yet Shafer himself (in press) has informally proposed a non-monotonic revision (discounting) of the input belief functions when significant conflict leads to counter-intuitive conclusions (Cohen, et al., 1985). The NMP system implements this revision process within an expert system. Belief functions are represented as based on assumptions: the system initially acts as if all discrediting factors are absent. When conflict arises, the system searches for a "culprit assumption" (the absence of a discrediting factor) and looks for evidence of its presence. When such evidence is found, the system modifies the culprit assumption by discounting the belief function associated with the discrediting factor. Unlike Doyle's system, the prioritization of the search for "culprits" is made explicit and is based on a benefit-cost tradeoff, and the information search is undertaken only if the possible result is worth its cost (otherwise, the system uses the across-the-board discounting of Step 4).

3. <u>Application to Air Force Threat Correlation and Route Replanning.</u> The NMP inference framework has been implemented in a small-scale prototype system designed to support pilots on deep interdiction or offensive counterair missions. The focus is in-flight route replanning in response to pop-up threats (i.e. threats which are discovered at sufficient range to permit rerouting). The system focuses on surface-to-air missile sites or artillery.

The Adaptive Route Replanner (ARR) is assumed to begin its mission with prior information (represented by a belief function) about the location of a particular surface-based anti-air threat. During flight, the system is notified of a second threat localization (from a SAR signal), represented by a second belief function. The system must assign degrees of belief among three possibilities: (1) the two belief functions represent the same threat, in the same location; (2) the original threat has moved to a new location; and (3) the second signal comes from an entirely new threat, previously undetected. The system must also provide measures of uncertainty regarding the locations of the threat(s).

In addition to the two belief functions representing threat localizations, ARR also has belief functions representing its domain knowledge, such as the range of distances a threat can move; how thorough was prior area intelligence (i.e. how likely is it that a threat was missed); and how far from the original threat is the second threat likely to be.

ARR's evidence is summarized by five belief functions over the space $A \times A \times T$, where A is the area in which the threats may be located (for simplicity, as-sumed to be 2-dimensional space), and T={S,D} is the set indicating whether the two signals represent the same or different threats. The two copies of A represent the two threat localizations. The three hypotheses of interest are: *same threat, unchanged location (U)*, represented by the set {(a,a,S): a∈A}; *threat has moved (M)*, represented by the set {(a,b,S): a,b∈A, a≠b}; *different threats (D)*, represented by the set {(a,b,D): a,b∈A}. (Note that the symbols a and b represent vectors in 2-dimensional space A.)

The five belief functions are described below.

- $Bel_1$: Summarizes prior evidence about the location of the first threat. Focal elements are of the form $S_x(a_1) \times A \times T$, where $S_x(a_1)$ is a circle of radius x centered at the point $a_1$. Thus, the evidence localizes the first threat with increasing belief within circles of increasing radius, but contains no information about the location of the second threat or whether they are the same or differnt.

- $Bel_2$: Summarizes evidence about the location of the second threat. Analogously to $Bel_1$, focal elements are of the form $A \times S_y(a_2) \times T$, localizing the threat within circles centered at $a_2$, but providing no information about the first threat or whether they are the same.

- $Bel_3$: Summarizes evidence about movement. One focal element is $(H \times \{S\}) \cup (A \times A \times \{D\})$, where $H = \{(a,a): a \in A\}$. This focal element represents belief in "unchanged, if same." The other focal elements are nested distance ranges, of the form $(C_w \times T) \cup (A \times A \times \{D\})$, where $C_w = \{(a,b): f_*(w) \le |a-b| \le f^*(w)\}$, with $f_*(w)$ and $f^*(w)$ being lower and upper bounds for the distance range of $C_w$. The $C_w$ represent belief in "if moved, then the distance range is $f_*(w)$ to $f^*(w)$." $Bel_3$ contains no information about separation if threats are different.

- $Bel_4$: Summarizes evidence about the thoroughness of prior area intelligence. There are two focal elements, $A \times A \times \{S\}$ and $A \times A \times T$. The first represents belief in intelligence sufficiently thorough that a threat was unlikely to be missed (i.e. threats are the same). The second represents uncommitted belief. There is no information about the location of either threat.

- $Bel_5$: Summarizes evidence about how far apart different threats are likely to be. Focal elements are of the form $(B_z \times \{D\}) \cup (A \times A \times \{S\})$, where $B_z = \{(a,b): |a-b| \ge g_*(z)\}$, with $g_*(z)$ being a lower bound for the distance range of $B_z$. The $B_z$ represent belief in "if different, then threats are separated by at least $g_*(z)$." $Bel_5$ contains no information about separation if the threats are the same.

The system operates on these belief functions in three passes. (I) *Forward chaining combination of belief functions using Dempster's Rule*. The process ends here if conflict is below a pre-set threshold. (II) *Prioritization and (possible) performance of tests*. This step may be executed several times, until conflict is below threshold (in which case the process ends) or until no test meets the benefit-cost criteria (in which case the next pass is executed). (III) *Across-the-board discounting of all arguments*. After performing its inferences, the system derives the action implications--it combines its threat classification (moved, unchanged, or different) with its knowledge of the danger contours associated with the threats, and selects the best of several candidate routes.

The forward chaining process is a straightforward application of Dempster's Rule. Belief in the three hypotheses can be obtained by taking the marginal of the combined belief function over the set T. Figure 2 shows the results of forward chaining in a numerical example. The input belief functions are given in the top part of the figure, and the probabilities of the three hypotheses (unchanged, moved, different) are shown below.

Input Belief Functions

$Bel_1$:   Center of Contours = (20, 20)

| Radius | 4.5 | 9.5 | 15.0 | 22.0 | 33.0 | 60.0 |
|---|---|---|---|---|---|---|
| Committed Belief | .18 | .18 | .18 | .18 | .18 | .10 |

$Bel_2$:   Center of Countours = (80, 80)

| Radius | 9.0 | 19.0 | 30.0 | 45.0 | 70.0 | 120.0 |
|---|---|---|---|---|---|---|
| Committed Belief | .18 | .18 | .18 | .18 | .18 | .10 |

$Bel_3$:   Belief Assigned to Diagonal = 0.3

| Lower Distance | 10.0 | 9.0 | 7.5 | 6.0 | 0.0 |
|---|---|---|---|---|---|
| Upper Distance | 13.0 | 15.0 | 18.0 | 22.0 | $\infty$ |
| Committed Belief | .15 | .15 | .15 | .15 | .10 |

$Bel_4$:   Belief Assigned to Same Threat = 0.7

$Bel_5$:

| Lower Distance | 60.0 | 49.0 | 40.0 | 32.0 | 26.0 | 20.0 |
|---|---|---|---|---|---|---|
| Upper Distance | $\infty$ | $\infty$ | $\infty$ | $\infty$ | $\infty$ | $\infty$ |
| Committed Belief | .17 | .17 | .17 | .17 | .17 | .15 |

Combined Belief Function: Classification of Second Threat
(U = unchanged; M = moved; D = different)

$Bel_*(\{U\})$ = .089     $Pl_*(\{U\})$ = .170
$Bel_*(\{M\})$ = .373     $Pl_*(\{M\})$ = .575
$Bel_*(\{D\})$ = .297     $Pl_*(\{D\})$ = .508

Conflict (Mass Assigned to Null Set) in Combined Belief Function = .409

Figure 2:  Example of Forward Chaining

In this example, conflict, or mass assigned to the null set, is a relatively large 0.42. It is instructive to consider how conflict arises. Conflict occurs to the extent that there is evidence *against* all three hypotheses (U, M, and D). Figure 3 shows six different types of conflict, corresponding to the six different ways in which mass can be assigned to the null set when Dempster's Rule is applied. The rows are the conflict types, and the columns represent *reasons* for the conflict. Thus, the first type of conflict occurs to the extent that belief is committed to non-overlapping contours of $Bel_1$ and $Bel_2$ (evidence against U), to non-movement (i.e. to the focal element $(H\times\{S\})\cup(A\times A\times\{D\})$ of $Bel_3$--evidence against M), and to distances incompatible with the distance contours of $Bel_5$ (evidence against D). The fifth type of conflict occurs to the extent that there is belief committed to distances incompatible with the movement contours of $Bel_3$ (evidence against M), but to movement in case threats are the same (i.e. not committed to the focal element $(H\times\{S\})\cup(A\times A\times\{D\})$--evidence against U), and to good prior intelligence coverage (evidence against D).

The conflict level of 0.42 exceeds the threshold for initiating the second pass of the inference mechanism (the threshold was set to 0.25 in our example). In the second pass, the system first decides on a discrediting factor to test for and a test to perform. The choice is based on a cost-benefit tradeoff. The *potential benefit* of a test is defined as the maximum possible conflict reduction that could occur as a result of performing the

163

test. Because conflict reduction is linear in the discount rate (Cohen, Laskey, and Ulvila, 1986), the benefit is the product of the derivative (with respect to conflict) of the belief function associated with the discrediting factor, and the maximum discount rate that could occur as a result of the test. The test is chosen for which the potential benefit per unit cost is the highest.

| CONFLICT TYPE | CONTOUR NON-OVERLAP | DISTANCE PRECLUDES MOVEMENT | EVIDENCE FOR MOVEMENT (if same threat) | EVIDENCE AGAINST MOVEMENT | DISTANCE PRECLUDES DIFFERENT | COVERAGE GOOD |
|---|---|---|---|---|---|---|
| (1) | X |   |   | X | X |   |
| (2) | X |   |   | X |   | X |
| (3) |   | X | X |   | X |   |
| (4) | X | X |   |   | X |   |
| (5) |   | X | X |   |   | X |
| (6) | X | X |   |   |   | X |

Figure 3: Reasons for Conflict--Six Types of Conflict

In our example, the system decides to test for the presence of ECM (electronic countermeasures) in the area. Performing the test results in discounting the belief function $Bel_2$ by a discount rate of 0.4. The new belief assignments are displayed in Figure 4. Two features are worthy of note. First, the amount by which the belief assignments for the three hypotheses fail to sum to 1.0 (a measure of incompleteness of evidence) increases after discounting (from 0.241 to 0.309). Second, belief in M and U has increased relative to belief in D (we discounted one of the belief functions contributing to evidence against U; our evidence against D, the belief in thorough intelligence, has not changed). Conflict has now been reduced to just below the threshold of 0.25. If conflict remained above the threshold, the system would search for another test to perform. If no acceptable test were found (i.e. no test for which the benefit/cost ratio was above a threshold), then the system would proceed to the overall discounting step.

ECM Present: $Bel_2$ discounted (discount rate = 0.4)

$Bel_*(\{U\}) = .153$  $Pl_*(\{U\}) = .292$
$Bel_*(\{M\}) = .398$  $Pl_*(\{M\}) = .641$
$Bel_*(\{D\}) = .140$  $Pl_*(\{D\}) = .398$

Figure 4: Results of Discounting: ABM Classification

Changes in the example have an impact agreeing with intuition. If the second threat localization is closer to the first threat (say, (50,60) instead of (80,80)), conflict in the initial pass is only .17, so the second pass is never initiated. Belief in a moved threat is fairly high ($Bel_*(\{U\}) = .15$; $Bel_*(\{M\}) = .43$; $Bel_*(\{D\}) = .09$). If instead, belief in thorough area intelligence is reduced to 0.3, conflict is again below threshold at .18, but this time the combined belief function favors different threats ($Bel_*(\{U\}) = .02$; $Bel_*(\{M\}) = .11$; $Bel_*(\{D\}) = .51$).

4. <u>Conclusions</u>. In virtually all problem solving domains where expert systems technology might be introduced, there is need for explicit and valid quantitative modeling of uncertainty. At the same time, we argue the need for a metastructure of qualitative reasoning in which the assumptions utilized in the probability model are reassessed and revised in the course of the argument. These dual requirements are addressed by the NMP inference framework.

The prototype system described above has demonstrated the feasibility of a self-revising inference engine for handling uncertainty, and has produced a preliminary design for expert systems inferencing with a very wide potential applicability.

# REFERENCES

Cohen, M.S., Schum, D.S., Freeling, A.N.S., and Chinnis, J.O., Jr. *On the art and science of hedging a conclusion: Alternative theories of uncertainty in intelligence analysis* (Technical Report 84-6). Falls Church, VA: Decision Science Consortium, Inc., August 1985.

Cohen, M.S., Laskey, K.B., and Ulvila, J.W. *Report on estimating uncertainty: Application to sample problems*. Falls Church, VA: Decision Science Consortium, Inc., March 1986.

Doyle, J. A truth maintenance system. *Artificial Intelligence*, 1979, *12(3)*, 231-272.

Shafer, G. *A mathematical theory of evidence*. Princeton, NJ: Princeton University Press, 1976.

Shafer, G. *Belief functions and possibility measures*. Lawrence, KS: University of Kansas, School of Business, in press.

Toulmin, S., Rieke, R., and Janik, A. *An introduction to reasoning* (2nd Edition). NY: Macmillan Publishing Company, 1984.

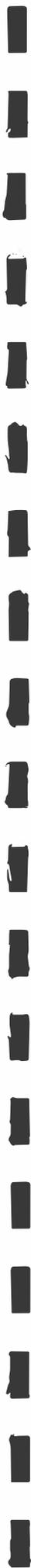